# COMPRESSING COMPLEX CONVOLUTIONAL NEURAL NETWORK BASED ON AN IMPROVED DEEP COMPRESSION ALGORITHM

*Jiasong Wu*[1, 2, 3, 4], *Hongshan Ren*[1, 4], *Youyong Kong*[1, 4], *Chunfeng Yang*[1, 4], *Lotfi Senhadji*[2, 3, 4], *Huazhong Shu*[1, 4]

[1]LIST, the Key Laboratory of Computer Network and Information Integration (Southeast University), Ministry of Education, 210096 Nanjing, China
[2]INSERM, U1099, Rennes, F-35000, France
[3]Université de Rennes 1, LTSI, Rennes, F-35042, France
[4]Centre de Recherche en Information Médicale Sino-français (CRIBs), Southeast University, Université de Rennes 1, INSERM

## ABSTRACT

Although convolutional neural network (CNN) has made great progress, large redundant parameters restrict its deployment on embedded devices, especially mobile devices. The recent compression works are focused on real-value convolutional neural network (Real CNN), however, to our knowledge, there is no attempt for the compression of complex-value convolutional neural network (Complex CNN). Compared with the real-valued network, the complex-value neural network is easier to optimize, generalize, and has better learning potential. This paper extends the commonly used deep compression algorithm from real domain to complex domain and proposes an improved deep compression algorithm for the compression of Complex CNN. The proposed algorithm compresses the network about 8 times on CIFAR-10 dataset with less than 3% accuracy loss. On the ImageNet dataset, our method compresses the model about 16 times and the accuracy loss is about 2% without retraining.

***Index Terms*—**Deep learning, complex-value convolutional neural network, compression, deep compression

## 1. INTRODUCTION

Convolutional neural networks (CNNs) have significantly improved the performances and the accuracy of image processing, speech processing, natural language processing and so on. CNNs have wide applications in the industry, such as the use of speech recognition of Baidu and Microsoft Cortana virtual assistant on speech recognition. Due to the huge amount of parameters and the computational burden of CNNs, the intelligence services are generally run in large professional equipment. Redundant parameters in the convolution neural network make it a highly intensive computing and memory intensive consuming model. Thus, the application of CNNs to embedded or mobile devices encounter three major difficulties: a) Huge model in terms of memory. For example, AlexNet [1] model is more than 200MB and VGG-16 [2] model is more than 500MB; b) Large amount of calculation. The well performing convolutional neural network (CNN) model has thousands of parameters, and it consumes a lot of time to run an intelligent service to get the result; c) Great power consumption. A large number of accesses to memory and uses of CPU resources will lead to huge power consumption.

With the popularization of mobile devices, the application demand of CNN in embedded systems is increasing. However, for equipment with limited hardware resources, the complete CNN model can hardly be directly transplanted to embedded devices. Compared with real-value convolutional neural network (Real CNN), complex-value convolutional neural network (Complex CNN) [3] has obvious parameter reduction. In addition, Complex CNN and its variants [3-10] achieve competitive performance in image classification and state-of-the-art results in music transcription tasks, etc. But the number of parameters in Complex CNN is still not small enough for embedded devices.

On the other hand, recent model compression works are concentrated on Real CNN, and to our knowledge, there is no attempt for the compression of Complex CNN. In the compression of Real CNN, many researchers have made effective contributions, among which "Deep Compression" [11] has achieved remarkable compression effect. Whereas for the Complex CNN, no one has been involved so far.

In this paper, we propose an improved deep compression algorithm for Complex CNN. Unlike Real CNN, we need to maintain the correlations between real and imaginary parts of complex weights in Complex CNN. Our major contributions are as follows:

a) Pruning: real numbers can be compared directly by their absolute values. In order to maintain correlations of real and imaginary parts, complex numbers need to be compared by their modules.

b) Quantization: the weight in real network is one-dimensional so one-dimensional K-means clustering method can be simply used. However, the weight in Complex CNN is two-dimensional, so it is necessary to apply two-dimensional K-means clustering to implement quantification. The centroid initialization of Complex CNN is different from Real CNN as well. Although both use linear initialization to achieve better results, Real CNN only distributes the centroid evenly between the maximum and minimum of the weights. For Complex CNN, the centroid linear initialization is more complicated and divided into four types as described in Section 3. We need to calculate the line according to the weight distribution in different schemes, and then take the centroid evenly on the line.

c) Huffman coding: Real CNN simply directly encodes the weights, while the Complex CNN needs to encode and store the real and complex parts of the weights separately.

The proposed deep compression algorithm was tested on CIFAR-10 and ImageNet datasets. Experiment results show that the proposed algorithm compresses the Complex CNN about 8 times on CIFAR-10 and 16 times on ImageNet, and the accuracy drops within 3% and 2%, respectively.

## 2. RELATED WORK

For the compression of Real CNN, researchers proposed many methods, which can be roughly divided into four classes: parameter sharing [12-16], network pruning [11, 17, 18], knowledge distillation [19] and matrix decomposition theory [20-24].

The main idea of parameter sharing is to share the same value with multiple parameters, and the actual implementation methods are also different. Vanhoucke and Mao [12] reduced the precision of parameters by means of fixed-point method, so that the parameters with similar values share the same weight. Chen et al. [13] proposed a method based on the hash algorithm to map the parameters to the corresponding hash bucket and share the same value in the same hash bucket. Gong et al. [14] used the K-means clustering algorithm to quantize the parameters, and the parameters of each cluster share its central value.

Network pruning can be used to reduce network complexity and effectively prevent overfitting. Han et al. [11] compressed the network by removing the network connection under a certain threshold to compress the trained network, and then further compressed the network based on parameter sharing and Huffman coding.

Knowledge distillation compresses the networks by transferring the knowledge of a large cumbersome model to a small simple one. Sau and Balasubramanian [19] used knowledge distillation method to compress networks based on the teacher-student network model and test on the MNIST dataset. The results showed that the method reduces simultaneously the storage and computational complexity of the model.

In the compression method based on the matrix decomposition theory, Denil et al. [20] and Nakkiran et al. [21] both used low rank decomposition to compress the parameters in the different layers of the neural network. Denton et al. [22] used the matrix decomposition method in the CNN to accelerate the calculation of the convolutional layer and effectively reduced the parameters of the fully connected layer.

These methods compress the network to different degrees, and all of them have achieved good results, but they all focus on Real CNNs, and to the best of our knowledge, there is no compression algorithm for Complex CNNs.

## 3. DEEP COMPRESSION MODEL FOR COMPLEX CNN

The deep compression algorithm for Real CNN, proposed by Han et al. [11], contains three stages as shown in Fig. 1 (top): weight pruning, weight quantization and Huffman coding. The algorithm can reduce the parameters of AlexNet [1] from 240MB to 6.9MB (the compression ratio is about 35), and the parameters of VGG-16 [2] decrease from 552MB to 11.3MB (the compression ratio is about 49) almost without loss of accuracy.

As shown in Fig. 1 (bottom), we proposed an improved deep compression method for Complex CNN and this method also includes three stages that are described in detail in the following.

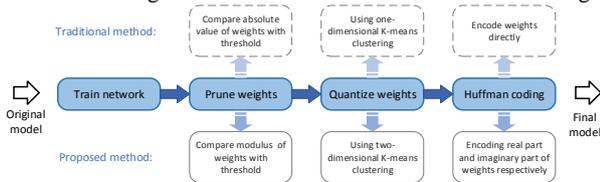

Fig. 1. Three stage of compression: pruning, quantization and Huffman coding

### 3.1. Weight pruning

Weight pruning is proved to be an effective way to reduce the redundancy of network weights. First, we train the baseline Complex CNN [3] to learn the connectivity of the network. Then, remove the "small" weight connections in the network. In Real CNN, we usually set a real number threshold and compare the absolute value of weights to the threshold to determine which are the "small" weights that need to be pruned. However, the weights in the Complex CNN [3] are complex numbers, and there is no direct comparison between two complex numbers. Here, three schemes can be chosen to solve the problem: compare the real part of the complex weight with the given real number threshold; compare the imaginary part of the complex weight with the given threshold; **compare the modulus of the complex weight and the given threshold.** From the experimental results, the best result is obtained from the complex module scheme, the reason maybe that the complex modulus contains both the real and the imaginary part information of the weights. Note that weight pruning allows obtaining a sparse network, and we can save the sparse matrix by using the method of compressed sparse row (CSR) or compressed sparse column (CSC).

### 3.2. Network quantization and weight sharing

Weight quantization further compresses pruned network by reducing the number of bits needed to represent each weight. At first, the weights that need to be stored each layer are represented by sharing the same weights; then, these weights are fine-tuned. For each connection layer, only shared weights and a small-shared weight index table are needed to be stored.

*3.2.1. Weight sharing*
The weight sharing can be solved by the K-means algorithm: using the K-mean clustering to identify the shared weights of each layer of the training network, so that the weights falling into the same cluster can be shared. Consequently, the number of parameters is reduced by storing only shared weights.

We expect to maintain the correlations between the real and the imaginary parts of a complex weight when dealing with the Complex CNN. **So we use the two-dimensional K-means clustering algorithm instead of one-dimensional one to carry out the clustering of complex weights as follows:**

$$\arg\min_{c} \sum_{i=1}^{m} \sum_{w \in c_i} \| w - c_i \|^2 \qquad (1)$$

where $w$ denotes complex weight, and each weight is in the form of $w_i = a_{w_i} + jb_{w_i}$, and $c_i$ denotes the centroid of each cluster and is in the form of $c_i = a_{c_i} + jb_{c_i}$.

We use Euclidean distance to measure the distance between two points, as formula (1), and partitioning $n$ original complex weights $W = \{w_1, w_2, ..., w_n\}$ into $m$ clusters $C = \{c_1, c_2, ..., c_m\}$, $m<<n$, to minimize the within-cluster sum of squares.

*3.2.2. Centroid initialization*
The centroid obtained by clustering is a shared weight. The centroid initialization affects the quality of clusters, thus impacting the prediction accuracy of the network. There are usually three ways to initialize the centroid: Forgy (random) initialization, density-based initialization and linear initialization.

**Since the weights of complex networks are two-dimensional, the centroid linear initialization of complex networks can be divided into the following four types:**

a) Linear initialization method in horizontal direction: the centroid is evenly distributed on the straight line $y=l$, where $l$ is a constant.

b) Linear initialization method in vertical direction: the centroid is evenly distributed on the straight line $x=l$, where $l$ is a constant.

c) Linear initialization method in positive inclined direction: the centroid is evenly distributed on the straight line $y=kx+v$, where $k$ is a positive number and $v$ is a constant.

d) Linear initialization method in negative inclined direction: the centroid is evenly distributed on the straight line $y=-kx+v$, where $k$ is a positive number and $v$ is a constant.

Forgy and density-based initializations select centroid based on the weights' distribution. This will lead to very few centroids having large absolute value and result in poor representation of these few large weights. However, linear initialization evenly distributes the centroids in the maximum and minimum interval of the weights, so this method may obtain more large weights than the first two methods. The centroid linear initialization of Real CNN simply distributes centroid between the maximum and minimum weights of one-dimension. In contrast, the linear initialization of Complex CNN requires a straight line computed in one of the four ways mentioned above, and then the centroid which is a binary tuple in complex value form is evenly taken on the straight line. As a result, **we choose linear initialization to do clustering experiments in section 4 and the results show that the fourth linear initialization method works best.**

### 3.3. Huffman coding

The Huffman approach encodes the source symbols by using a variable length coding table, which is obtained by evaluating the probability of the occurrence of a source symbol, and the higher frequency character uses a shorter encoding, and vice versa. After that, the average length and expected value of the encoding string are reduced, so as to achieve the goal of lossless compression of data.

The storage of the complex weights of Complex CNN is in the form of a pair of real numbers. Huffman coding is based on the frequency of character occurrence. The coding result has nothing to do with the correlations between the real part and the imaginary part of the weight. Therefore, **the real and imaginary parts can be respectively coded to get the final compression network.**

## 4. EXPERIMENTAL RESULTS

We first train benchmark Complex CNN [3] shown in Fig. 2, and then we perform compression. Network pruning is implemented by forcing the weight below the threshold to be zero. In the quantification stage, we used different quantities of clustering at different network stages. Finally, we apply Huffman coding on the real part and the imaginary part of the complex weights, respectively.

We give the experiment of the compression of Complex CNN on two datasets: CIFAR-10 and IMAGENET. All the following experiments are performed on the Tensorflow backend of the Keras framework on the system Ubuntu 16.04.

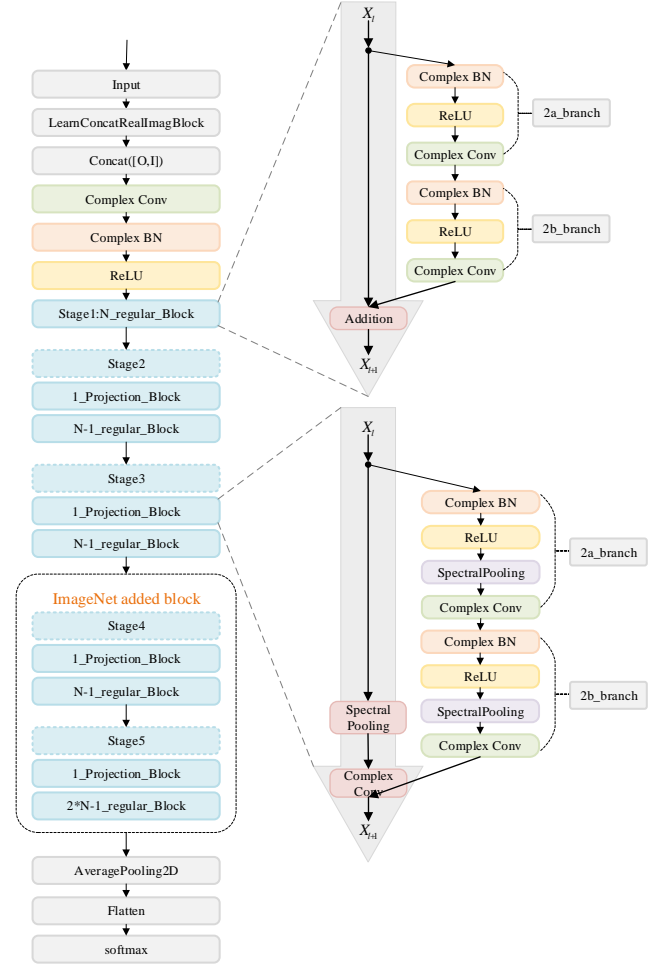

Fig. 2. Two different scales of network architecture trained on CIFAR-10 and ImageNet. Left is the layer composition of the whole model, including convolution layer, BN layer, residual block, etc. The difference of the model between the CIFAR-10 and ImageNet datasets is that the former only contains stages 1-3 and the latter contains stages 1-5. Right is the composition of the projection block and regular block. The projection block has one more spectral pooling layer before complex convolution layer than the regular block in every branch.

### 4.1. CIFAR-10

We trained CIFAR-10 dataset on the baseline Complex CNN [3] as shown Fig. 2. When no retraining is performed at the pruning and quantization stages and the related parameters are not optimized, the initial results of the compressed Complex CNN on the CIFAR-10 are shown in Tables 1 and 2. As shown in Table 1, with the increase of threshold value, the pruning ratio increases and the accuracy rate decreases. At the threshold of 0.03, the two indicators reach a compromise. As shown in Table 2, on the whole, the number of clusters and the accuracy demonstrate a normal distribution. The results on CIFAR-10 show that the compression ratio of compressed Complex CNN can reach 8 with less than 3% accuracy loss. Among them, most of the saving comes from quantization. Since the original network model is smaller than other large Real CNN [1, 2], pruning and Huffman coding contribute a little.

Table 1. Accuracy and pruning ratio in pruned Complex CNN on CIFAR-10.

| Threshold | Accuracy (%) | Pruning ratio (%) |
|---|---|---|
| Original Complex CNN | 93.19 | - |
| 0.01 | 93.14 | 10.9 |
| 0.02 | 93.07 | 25.49 |
| **0.03** | **92.98** | **38.7** |
| 0.04 | 92.35 | 50.86 |
| 0.05 | 89.65 | 61.38 |

Table 2. Accuracy and compression ratio in pruning, quantization and Huffman coding on CIFAR-10.

| Cluster number | | | Accuracy (%) | Storage size | Compress ratio |
|---|---|---|---|---|---|
| stage1 | stage2 | stage3 | | | |
| Original Complex CNN | | | 93.19 | 4.1MB | - |
| Pruned Complex CNN | | | 92.98 | 3.5MB | 1.2 |
| 100 | 100 | 100 | 90.74 | 557.2KB | 7.36 |
| 100 | 120 | 120 | 90.67 | 557.8KB | 7.35 |
| **90** | **100** | **100** | **90.72** | **552.2KB** | **7.42** |
| 80 | 150 | 150 | 90.17 | 547.1KB | 7.49 |
| Huffman coding | | | - | 502.4KB | 8.16 |

**4.2. IMAGENET**

In order to get better results on ImageNet dataset, we have added two stages to the baseline Complex CNN [3] as shown in Fig. 2 and adjusted the learning rate of model in different epochs. We start the learning rate of model at 0.1 for the first 10 epochs to speed up learning features and then set it to 0.01 from epoch 10-170 and finally set it to 0.001 in the rest of epochs. We set the kernel of the first convolution layer from $3\times3$ to $7\times7$ to learn more features.

When no retraining is performed at the pruning and quantization stages and the related parameters are not optimized, the initial results of the compressed Complex CNN on the ImageNet dataset are shown in Tables 3 and 4. As shown in Table 3, the pruning rate grows with the increase of threshold value and the Top-1 and Top-5 accuracy both point out a fluctuating situation. At the threshold of 0.009, the indicators reach a compromise. In the stage of quantization, the number of clusters of stages 3-5 must be set to 256, and if changed to other values, the result of clustering will be unsatisfactory. The number of cluster of stage 1-2 can be set near to 127 or 256 and the best result appears when setting the cluster's number of stage 1-2 to 127 and stage 3-5 to 256. As shown in Table 4, network pruning contributes half of the compression rate. Quantification contributes 7 times more to compression rate than the former and Huffman coding is less effective here. As a result, the compression ratio of compressed Complex CNN can reach 16 with less than 2% both top-1 accuracy and top-5 accuracy loss.

Table 3. Accuracy and pruning ratio in pruned Complex CNN on ImageNet.

| Threshold | Top-1 accuracy (%) | Top-5 accuracy (%) | Pruning ratio (%) |
|---|---|---|---|
| Original Complex CNN | 68.31 | 88.07 | - |
| 0.006 | 67.67 | 88.01 | 19.13 |
| 0.007 | 67.61 | 87.94 | 24.56 |
| 0.008 | 67.41 | 87.77 | 29.48 |
| **0.009** | **68.10** | **88.04** | **34.13** |
| 0.01 | 67.38 | 88.05 | 38.39 |

Table 4. Accuracy and compression ratio in pruning, quantization and Huffman coding on ImageNet.

| Cluster number | | | Top-1 Accuracy (%) | Top-5 Accuracy (%) | Storage size | Compress ratio |
|---|---|---|---|---|---|---|
| stage1/2 | stage3 | stage4/5 | | | | |
| Original Complex CNN | | | 68.31 | 88.07 | 51.6MB | - |
| Pruned Complex CNN | | | 68.10 | 88.04 | 25.2MB | 2.05 |
| 126 | 256 | 256 | 63.99 | 85.74 | 3.6MB | 14.33 |
| **127** | **256** | **256** | **66.11** | **86.73** | **3.6MB** | **14.33** |
| 128 | 256 | 256 | 64.81 | 86.25 | 3.6MB | 14.33 |
| 130 | 256 | 256 | 42.08 | 68.09 | 3.6MB | 14.33 |
| 256 | 256 | 256 | 65.83 | 86.39 | 3.6MB | 14.33 |
| Huffman coding | | | - | - | 3.2MB | 16.12 |

**5. CONCLUSION**

In this paper, we first proposed an improved deep compression algorithm for Complex CNN. In the pruning stage, we use the modulus instead of absolute value of complex number for thresholding. In the quantization stage, we use two-dimensional K-means clustering instead of one-dimensional to cluster the complex weights. Finally, the Huffman code is used to further compress the parameters and to encode both of the real and imaginary parts of complex weights. We have conducted experiments on CIFAR-10 and ImageNet and adjusted the structure of Complex CNN to accommodate larger scale datasets. The results show that the compression rate is about 8 on CIFAR-10 and 16 on ImageNet, and the accuracy loss is within 3% and 2% respectively. Note that if we retrain the compressed model, the accuracy will be improved somewhat.

In the future, it will be interesting to focus on efficient operations to speed up compressed models. We expect to combine our model with other network acceleration algorithms, such as Low Rank algorithm [24] and Winograd convolution algorithms [25, 26] to further reducing computational costs.

**ACKNOWLEDGEMENT**

This work was supported in part by the National Natural Science Foundation of China under Grants 61876037, 31800825, 61871117, 61871124, 61773117, 31571001, 61572258, and in part by the National Key Research and Development Program of China under Grant 2017YFC0107900, and in part by the Short-Term Recruitment Program of Foreign Experts under Grant WQ20163200398.